\documentclass[conference,onecolumn, 11pt]{IEEEtran}
\IEEEoverridecommandlockouts
\usepackage{siunitx}
\usepackage[acronym]{glossaries}
\usepackage{ifthen}
\usepackage{amsmath,amssymb,amsfonts}
\usepackage{algorithmic}
\usepackage{graphicx}
\usepackage{textcomp}
\usepackage{xcolor}
\usepackage{colortbl}
\usepackage{ragged2e}
\usepackage{tabularx}
\usepackage[bookmarks=false]{hyperref}
\usepackage{comment}
\usepackage{booktabs}
\usepackage{multirow}
\usepackage{dirtytalk}
\usepackage{tikz}
\usetikzlibrary{shapes.geometric, arrows, positioning}
\usepackage{circuitikz}
\usepackage{svg}
\usepackage{pgfplots}
\usepgfplotslibrary{fillbetween}
\pgfplotsset{width=10cm,compat=1.9}
\usepackage[most]{tcolorbox}
\usepackage{enumitem}
\usepackage{caption}
\usepackage{listings}

\usepackage{titlesec}

\definecolor{ForestGreen}{RGB}{34,150,34}

\newboolean{blindreview}
\setboolean{blindreview}{false}

\newacronym{SoC}{SoC}{System-on-Chip}
\newacronym{AI}{AI}{Artificial Intelligence}
\newacronym{ML}{ML}{Machine Learning}
\newacronym{LLMs}{LLMs}{Large Language Models}
\newacronym{LLM}{LLM}{Large Language Model}
\newacronym{ASIC}{ASIC}{Application Specific Integrated Circuit}
\newacronym{DUV}{DUV}{Design Under Verification}
\newacronym{CEX}{CEX}{Counter Example}
\newacronym{FSM}{FSM}{Finite State Machine}
\newacronym{FSMs}{FSMs}{Finite State Machines}
\newacronym{RTL}{RTL}{Register Transfer Level}
\newacronym{IP}{IP}{Intellectual Property}
\newacronym{NLP}{NLP}{Natural Language Processing}
\newacronym{HDL}{HDL}{Hardware Description Language}
\newacronym{GPT}{GPT}{Generative Pre-trained Transformer}
\newacronym{BERT}{BERT}{Bidirectional Encoder Representations from Transformers}
\newacronym{PaLM}{PaLM}{Pathways Language Model}
\newacronym{LLaMA}{LLaMA}{Large Language Model Meta AI}
\newacronym{SVA}{SVA}{SystemVerilog Assertion}
\newacronym{SVAs}{SVAs}{SystemVerilog Assertions}
\newacronym{FV}{FV}{Formal Verification}
\newacronym{GenAI}{GenAI}{Generative Artificial Intelligence}
\newacronym{BMC}{BMC}{Bounded Model Checking}

\usepackage[style=ieee, backend=biber, natbib=true, maxbibnames=1, minbibnames=1]{biblatex}
\usepackage{fancyhdr}

\fancypagestyle{firstpage}{%
	\fancyhf{}%
	\fancyhead[C]{\normalsize To appear at the 37th IEEE International System-on-Chip Conference, Sep 16-19 2024, Dresden, Germany}
	
	\fancyfoot[C]{\footnotesize \copyright2024 IEEE. Personal use of this material is permitted. Permission from IEEE must be obtained for all other uses, in any current or future media, including reprinting/republishing this material for advertising or promotional purposes, creating new collective works, for resale or redistribution to servers or lists, or reuse of any copyrighted component of this work in other works.}
}

\begin{document}

\lstset{
    language=Verilog,           
    basicstyle=\footnotesize,   
    numbers=left,               
    frame=lines,                
    captionpos=b,               
    breaklines=true,            
    tabsize=2,                  
    xleftmargin=2.1em,
    framexleftmargin=1.7em,
    commentstyle=\color{ForestGreen},
    keywordstyle=\color{blue},
    stringstyle=\color{red},
}

\lstdefinelanguage{Verilog}{morekeywords={accept_on,alias,always,always_comb,always_ff,always_latch,and,assert,assign,assume,automatic,before,begin,bind,bins,binsof,bit,break,buf,bufif0,bufif1,byte,case,casex,casez,cell,chandle,checker,class,clocking,cmos,config,const,constraint,context,continue,cover,covergroup,coverpoint,cross,deassign,default,defparam,design,disable,dist,do,edge,else,end,endcase,endchecker,endclass,endclocking,endconfig,endfunction,endgenerate,endgroup,endinterface,endmodule,endpackage,endprimitive,endprogram,endproperty,endspecify,endsequence,endtable,endtask,enum,event,eventually,expect,export,extends,extern,final,first_match,for,force,foreach,forever,fork,forkjoin,function,generate,genvar,global,highz0,highz1,if,iff,ifnone,ignore_bins,illegal_bins,implements,implies,import,incdir,include,initial,inout,input,inside,instance,int,integer,interconnect,interface,intersect,join,join_any,join_none,large,let,liblist,library,local,localparam,logic,longint,macromodule,matches,medium,modport,module,nand,negedge,nettype,new,nexttime,nmos,nor,noshowcancelled,not,notif0,notif1,null,or,output,package,packed,parameter,pmos,posedge,primitive,priority,program,property,protected,pull0,pull1,pulldown,pullup,pulsestyle_ondetect,pulsestyle_onevent,pure,rand,randc,randcase,randsequence,rcmos,real,realtime,ref,reg,reject_on,release,repeat,restrict,return,rnmos,rpmos,rtran,rtranif0,rtranif1,s_always,s_eventually,s_nexttime,s_until,s_until_with,scalared,sequence,shortint,shortreal,showcancelled,signed,small,soft,solve,specify,specparam,static,string,strong,strong0,strong1,struct,super,supply0,supply1,sync_accept_on,sync_reject_on,table,tagged,task,this,throughout,time,timeprecision,timeunit,tran,tranif0,tranif1,tri,tri0,tri1,triand,trior,trireg,type,typedef,union,unique,unique0,unsigned,until,until_with,untyped,use,uwire,var,vectored,virtual,void,wait,wait_order,wand,weak,weak0,weak1,while,wildcard,wire,with,within,wor,xnor,xor,`uvm_create, `uvm_rand_send_with},morecomment=[l]{//}}

\title{Generative AI Augmented Induction-based Formal Verification\\
\thanks{This work has been developed in the project VE-VIDES (project label 16ME0243K) which is partly funded within the Research Programme ICT 2020 by the German Federal Ministry of Education and Research (BMBF).}
}

\ifthenelse{\boolean{blindreview}}{}{
\author{\IEEEauthorblockN{Aman Kumar}
\IEEEauthorblockA{\textit{Infineon Technologies} \\
Dresden, Germany \\
Aman.Kumar@infineon.com}
\and
\IEEEauthorblockN{Deepak Narayan Gadde}
\IEEEauthorblockA{\textit{Infineon Technologies} \\
Dresden, Germany \\
Deepak.Gadde@infineon.com}
}
}

\maketitle

\thispagestyle{firstpage}

\begin{abstract}
\textbf{\acrfull{GenAI} has demonstrated its capabilities in the present world that reduce human effort significantly. It utilizes deep learning techniques to create original and realistic content in terms of text, images, code, music, and video. Researchers have also shown the capabilities of modern \acrfull{LLMs} used by \acrshort{GenAI} models that can be used to aid hardware development. Formal verification is a mathematical-based proof method used to exhaustively verify the correctness of a design. In this paper, we demonstrate how \acrshort{GenAI} can be used in induction-based formal verification to increase the verification throughput.}
\end{abstract}


\section{Introduction}

Formal verification-based hardware verification is a highly desirable but labour-intensive task. It is an exhaustive verification technique that uses mathematical proof methods to verify if the design implementation matches its specifications \cite{aman_dvcon_ecc}. Although formal verification ensures the functional correctness of the design, achieving a full proof of properties written to verify the design could be challenging, especially for complex designs \cite{aman_dvcon_ecc}. Modern formal verification tools use \textit{k}-induction-based proof. This approach allows them to prove the design correctness for all time, rather than just the first few time steps \cite{cadence}. To avoid inductive step failures, it is necessary to use helper assertions that act as assumptions or invariants once proven. These helper assertions are also known as lemma. Moreover, in case an assertion fails in its inductive step, it takes human effort to find the root cause from \acrfull{CEX} and write a helper assertion to fix the failure. To address this complexity of formal verification, we explore the potential of \acrshort{GenAI} to automate assertion generation, which has proved in recent times to be a suitable tool \cite{llm_applications} \cite{assert_llm}. We demonstrate \acrshort{GenAI} based flow that can generate helper assertions based on a given \acrshort{RTL} design and generate lemmas in the event of an inductive step failure during formal verification.

\section{Background} \label{bgnd}

\subsection{Induction-based Formal Verification}

\acrfull{BMC} can find bugs in large designs. However, the correctness of a property is guaranteed only for the analysis bound. Induction-based proof must be applied to prove the design will work all the time rather than just the first \textit{S} time steps. Induction is a method to check if the design is in a random good state (state of design that doesn't produce a \acrshort{CEX}) whether it will be in a good state at the next cycle \cite{cadence}. Induction with increasing depth \textit{k}, consists of two steps, base and \textit{k}-step induction as described in \cite{induction_book}. The base case utilizes the initial-state constraint, whereas the inductive step does not. Consequently, the inductive step may encompass unreachable states. At the inductive stage, the formal tool unrolls the design into small pieces and is unaware of the unreachable states. It can start from any arbitrary (unreachable) state and end up in a state where the property fails \cite{sst_jug}. In practice, this might prevent the induction proof from succeeding without incorporating additional constraints, such as stronger induction invariants than the property itself \cite{induction_book}. These induction invariants or helper assertions rule out the \acrshort{CEX} by reducing the state-space during the inductive step.

\subsection{\acrfull{GenAI}}

\acrshort{GenAI} is a type of \acrshort{AI} technique that has the ability to produce original content across various formats, such as text, images, audio, and other forms of media. Propelled by the progress in deep learning, \acrfull{NLP}, and continuously improving data processing capabilities, \acrshort{LLMs} stand at the forefront of \acrshort{GenAI}. An \acrshort{LLM} is a type of \acrshort{GenAI} that specializes in understanding and generating human language by predicting subsequent parts of text based on the context provided in the input it receives. It is pre-trained on a large corpus of text and then fine-tuned for specific tasks or to improve its performance in generating coherent and contextually relevant text \cite{llm_book}. In hardware verification, \acrshort{GenAI} can be utilized to aid the verification processes such as assertion generation, failure debugging, testbench and stimulus generation \cite{llm_applications}.

\section{Lemmas Using Generative \acrshort{AI}} \label{lemma}

\begin{figure}[h!]
\centering
  \includegraphics [width=\columnwidth] {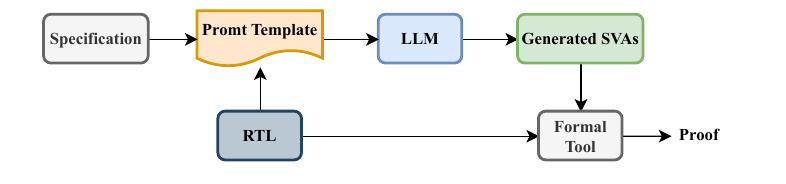}
\caption{Helper assertion generation using \acrshort{LLM}}
\label{helper_assertion}
\end{figure}

Fig.~\ref{helper_assertion} shows the flow to generate lemmas or helper assertions. There are two inputs to the \acrshort{LLM}. The first input is the specification document that contains information about the functionality of the design. The second input is the actual \acrshort{RTL} code written in any \acrfull{HDL}. The \acrshort{LLM} uses these two pieces of information and analyzes them to come up with a set of helper assertions that can be used for formal verification. In the next step, the generated assertions and the \acrshort{RTL} design are provided to the formal tool and the tool proves the correctness of the properties. Once proven, these assertions would be used as assumptions that help to prove other complex assertions.

\section{Induction Step Failure Analysis using Generative \acrshort{AI}} \label{induction}

\begin{figure}[h!]
\centering
  \includegraphics [width=\columnwidth] {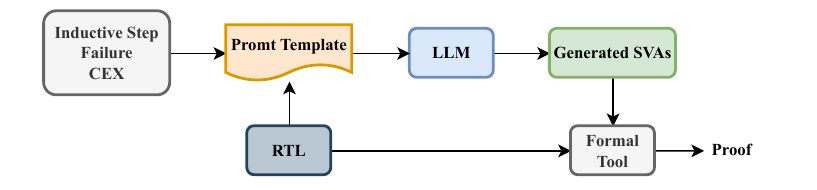}
\caption{Helper assertion generation for induction step failure using \acrshort{LLM}}
\label{induction}
\end{figure}

Fig.~\ref{induction} shows the flow to analyze the induction step failure during formal verification and use the capabilities of \acrshort{LLM} to generate helper assertions. The \acrshort{LLM} model requires two inputs. The first input is the \acrshort{CEX} from the inductive step failure. This is usually a waveform diagram that highlights the failure. The second input is the \acrshort{RTL} design that is used by the \acrshort{LLM} to analyze the design and relate it to the counterexample. After analysis, the \acrshort{LLM} generates the necessary helper assertion that prevents inductive step failure during the next iteration of formal proof. As an example, consider the simple design of two synchronized counters given in Listing~\ref{simple_design}.

\begin{lstlisting}[basicstyle=\small, language=Verilog, caption=Two synchronized counters design, label=simple_design]
module sync_counters (input clk, rst, output logic [31:0] count1, count2); 
  always @(posedge clk or posedge rst) begin 
    if (rst) begin 
      count1 <= 32'b0; 
      count2 <= 32'b0; 
    end else begin 
      count1++; 
      count2++; 
    end
  end
endmodule
\end{lstlisting}

\begin{figure}[h!]
    \centering
    \includegraphics[width=0.6\columnwidth]{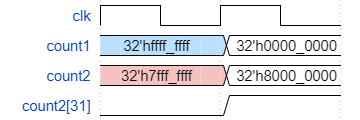}
    \caption{Induction step failure (31\textsuperscript{st} bit of count2 is not logic 1)}
    \label{cex}
\end{figure}

The following \acrshort{SVA} property in Listing~\ref{svacounter} should obviously hold true:

\begin{lstlisting}[basicstyle=\small, language=Verilog, caption=Two synchronized counters design, label=svacounter]
property equal_count;
  &count1 |-> &count2;
endproperty
\end{lstlisting}
However, the induction step fails with the \acrshort{CEX} in Fig.~\ref{cex}.

Upon providing the \acrshort{RTL} and the \acrshort{CEX}, the \acrshort{LLM} generated a helper assertion mentioned in Listing~\ref{sva} that proved the original assertion faster.

\begin{lstlisting}[basicstyle=\small, language=Verilog, caption=Helper assertion generated for the synchronized counters example, label={sva}]
property helper;
  count1 == count2;
endproperty
\end{lstlisting}

\section{Results} \label{results}

We used the helper assertion generation flow described in this paper for fairly complex designs. We utilized both flows to generate general helper assertions as well as for induction step failure. The designs used were counters and ECC. Usually, the flow was able to figure out necessary helper assertions that helped in faster proof for complex properties. It was also observed that the quality of generated assertions was much better in the case of LLMs from OpenAI such as GPT-4-Turbo and GPT-4o compared to Llama or Gemini. This could be due to the fact that the \acrshort{LLMs} from OpenAI are usually trained using relatively higher training data than the others.

\section{Conclusion} \label{conclusion}

In this paper, we have demonstrated a flow to generate helper assertions or lemmas using \acrshort{GenAI} to reduce the efforts for induction-based formal verification. The flow aims at generating usual helper assertions that reduce the proof time for complex properties as well as inductive invariants that help resolve induction step failures. Although LLMs prove to be effective in generating assertions for our designs, one must be aware of the limitations of using \acrshort{GenAI} especially for artificial hallucinations that produce vulnerable results \cite{reformai}. It is recommended to analyze the output from the \acrshort{LLM} before using it productively to avoid false positives and comprehend the so-called human-in-the-loop \acrshort{AI}.

\printbibliography[heading=bibintoc]

\end{document}